%% file: root.tex
\documentclass[letterpaper, 10 pt, conference]{IEEEtran}
\IEEEoverridecommandlockouts

\input{config}
\begin{document}

\title{\vspace{0.5cm}The Marathon 2:\\
A Navigation System}

\author{
    \IEEEauthorblockN{Steve Macenski}
    \IEEEauthorblockA{\textit{R\&D Innovations} \\
    \textit{Samsung Research}\\
    s.macenski@samsung.com}

    \and

    \IEEEauthorblockN{Francisco Mart\'in}
    \IEEEauthorblockA{\textit{Intelligent Robotics Lab} \\
    \textit{Rey Juan Carlos University}\\
    francisco.rico@urjc.es}

    \and

    \IEEEauthorblockN{Ruffin White}
    \IEEEauthorblockA{\textit{Contextual Robotics Institute} \\
    \textit{UC San Diego}\\
    rwhitema@eng.ucsd.edu}
    
    \and

    \IEEEauthorblockN{Jonatan Ginés Clavero}
    \IEEEauthorblockA{\textit{Intelligent Robotics Lab} \\
    \textit{Rey Juan Carlos University}\\
    jonatan.gines@urjc.es}
}

\maketitle
\begin{abstract}
Developments in mobile robot navigation have enabled robots to operate in warehouses, retail stores, and on sidewalks around pedestrians.
Various navigation solutions have been proposed, though few as widely adopted as ROS (Robot Operating System) Navigation.
10 years on, it is still one of the most popular navigation solutions\footnote{By GitHub stars, forks, citations, and industry use}.
Yet, ROS Navigation has failed to keep up with modern trends. We propose the new navigation solution, \textit{Navigation2}, which builds on the successful legacy of ROS Navigation.
Navigation2 uses a behavior tree for navigator task orchestration and employs new methods designed for dynamic environments applicable to a wider variety of modern sensors.
It is built on top of ROS2, a secure message passing framework suitable for safety critical applications and program lifecycle management.
We present experiments in a campus setting utilizing Navigation2 to operate safely alongside students over a marathon as an extension of the experiment proposed in Eppstein et al.\cite{eppstein}.
The Navigation2 system is freely available at \url{https://github.com/ros-planning/navigation2} with a rich community and instructions.
\end{abstract}

\begin{IEEEkeywords}
Service Robots; Software, Middleware and Programming Environments; Behaviour-Based Systems
\end{IEEEkeywords}

\input{body/introduction}
\input{body/related_work}
\input{body/design}
\input{body/implementation}

\input{body/analysis}
\input{body/limitations}
\input{body/conclusion}

\section*{Acknowledgements}
Other contributors on \textit{Navigation2} include: Matt Hansen, Carl Delsey, Brian Wilcox, Mohammad Haghighipanah, Mike Jeronimo, and Carlos Orduno. Authors also thanks Lorena Bajo for her work during experiments.


\end{document}

%% file: config.tex
\usepackage{cite}
\usepackage{amsmath,amssymb,amsfonts}
\usepackage{algorithmic}
\usepackage{graphicx}
\usepackage{textcomp}
\usepackage{hyperref}
\usepackage{xcolor}
\usepackage{subfig}
\usepackage{multirow}
\usepackage{soul}

\def\BibTeX{{\rm B\kern-.05em{\sc i\kern-.025em b}\kern-.08em
    T\kern-.1667em\lower.7ex\hbox{E}\kern-.125emX}}

%% file: body/introduction.tex
\section{Introduction}
\label{sec:introduction}

Many mobile robot navigation frameworks and systems have been proposed since the first service robots were created \cite{Burgard98} \cite{nav_survey}.
These frameworks laid the groundwork for the service robots that are being rolled out in factories, retail stores, and on sidewalks today.
While many have been proposed, very few have rivaled the impact of the \textit{ROS Navigation Stack} on the growing mobile robotics industry.

Indeed, this open-source project has fueled companies, governments, and researchers alike.
It has gathered more than 400 citations, over 1,000 papers indexed by Google Scholar mention it, and 1,200 forks on GitHub, as of February 2020.
Proposed in 2010, it provided a rich and configurable navigation system that has been reconfigured for use on a variety of robot platforms \cite{eppstein}.
It provided a reference framework and a set of foundational implementations of algorithms including A* \cite{astar} and Dynamic Window Approach (DWA) \cite{Fox97} for planning and control.
Navigation, however, suffered from perception and controller issues in highly dynamic environments and lacked a general state estimator.
Further, ROS1 lacks in security and performance aspects, rendering it unsuitable for safety critical applications. 

\input{figs/images/robots}

In this work, we studied current trends and robotics systems to create a new navigation system leveraging our collective experiences working with popular robotics frameworks.
This work looks to build off of the success of Navigation while supporting a wider variety of robot shapes and locomotion types in more complex environments.

In this paper, we propose a new, fully open-source, navigation system with substantial structural and algorithmic refreshes, \textit{Navigation2}.
Navigation2 uses a configurable behavior tree to orchestrate the planning, control, and recovery tasks \cite{bt_intro}.
Its federated model is structured such that each behavior tree node invokes a remote server to compute one of the above tasks with one of a growing number of algorithm implementations.
Each server implements a standard plugin interface to allow for new algorithms or techniques to be easily created and selected at run-time.

This architecture makes use of multi-core processors and leverages the real-time, low-latency capabilities of ROS2 \cite{ros2_performance}.
ROS2 was redesigned from the ground up to be industrial-grade providing security, reliability, and real-time capabilities for the next generation of robots.
ROS2 also introduces the concept of \textit{Managed Nodes}, servers whose life-cycle state can be controlled, which we exploit to create deterministic behavior for each server in the system.

Our proposed algorithmic refreshes are focused on modularity and operating smoothly in dynamic environments.
This includes: \textit{Spatio-Temporal Voxel Layer} (STVL) \cite{stvl}, layered costmaps \cite{dlu}, \textit{Timed Elastic Band} (TEB) controller \cite{teb}, and a multi-sensor fusion framework for state estimation, \textit{Robot Localization} \cite{rl}.
Each supports holonomic and non-holonomic robot types.
However, Navigation2 is specifically designed such that \textit{all} components are run-time configurable and in many cases, multiple alternative algorithms are already available for use.
Our navigation system, the first of its kind built using ROS2, is shown to safely operate in a highly-dynamic campus setting.
Through long-duration experiments, we show that it can robustly navigate a distance in excess of a marathon (37.4 miles total) with the Tiago \& RB-1 base, shown in Figure \ref{fig:robots}, through high-trafficked areas.

%% file: figs/images/robots.tex
\begin{figure}
    \centering
    \subfloat[Tiago \cite{tiago_img}]{
        \includegraphics[
            width=0.2\textwidth]
            {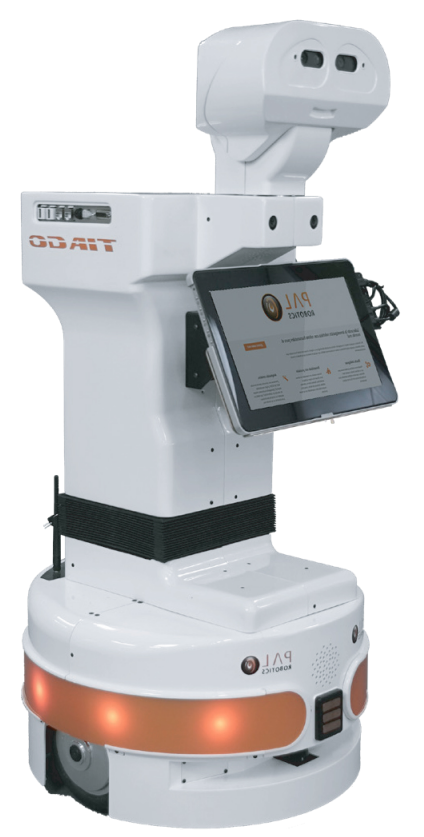}}%
    \subfloat[RB-1 \cite{rb1_img}]{
        \includegraphics[
            width=0.17\textwidth]
            {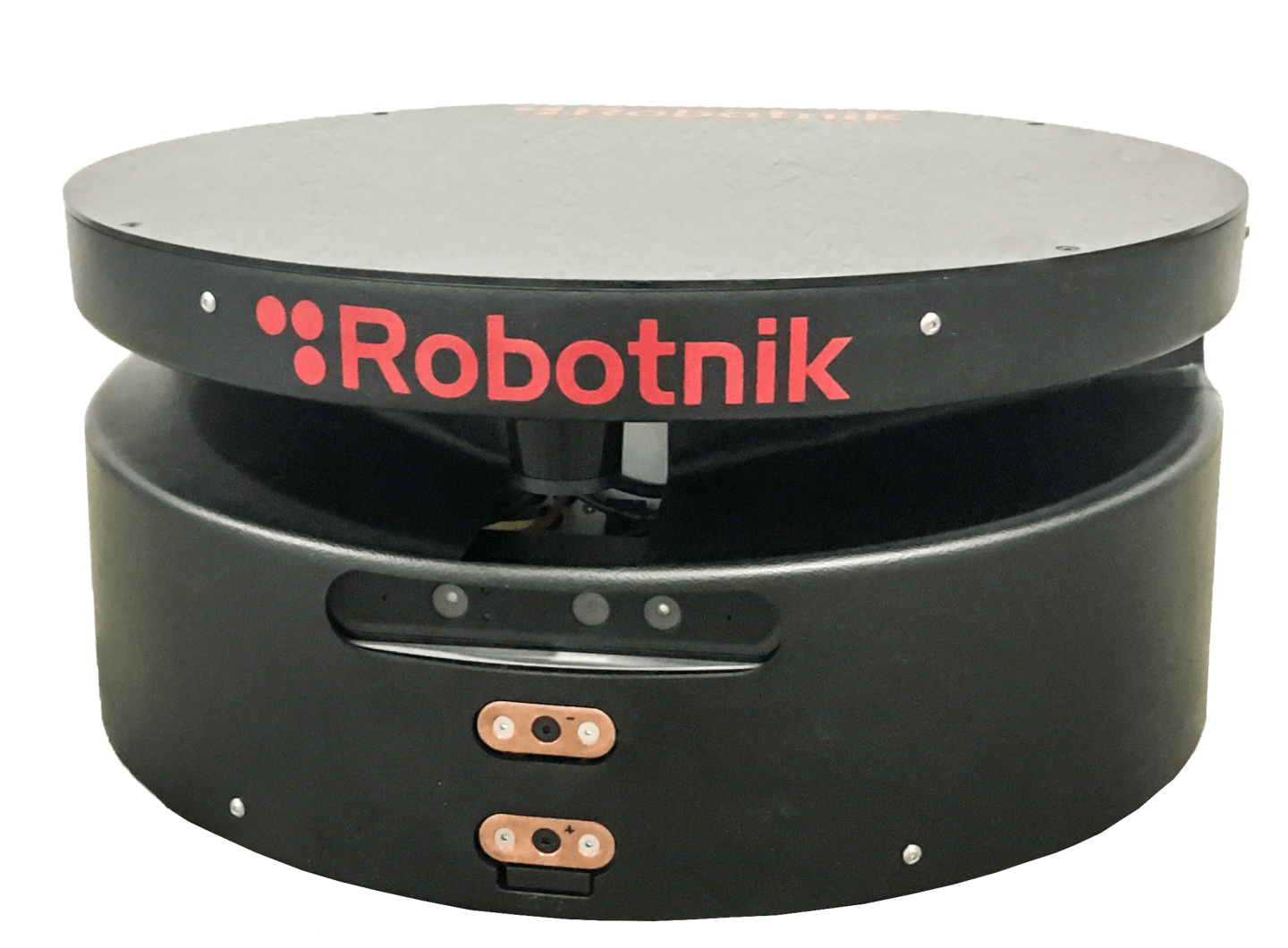}}
    \caption{Robots used for the marathon experiments.}
    \label{fig:robots}
\end{figure}

%% file: body/related_work.tex
\section{Related Work}
\label{sec:related_work}

\subsection{Navigation Systems}
Dervish was one of the first robots that implemented an effective navigation system in an office environment \cite{Nourbakhsh95}.
It used a coarse probabilistic representation of the environment using a sonar ring to navigate through a hallway scene for the Office Delivery Robot Competition. 
Soon after, \cite{Burgard98} and \cite{Thrun99} introduced the museum robots, RHINO and MINERVA.
These robots primarily used probabilistic occupancy grids to localize in their environments with varying planning approaches.
To avoid obstacles, DWA was used \cite{Fox97}.
These robots relied on 2D environmental representations for positioning and navigation, which could lead to collisions with out of plane obstacles.

\cite{eppstein} presented a navigation system and environmental representation utilizing 3D information from tilting 2D laser scanners to navigate an environment.
Many of the early techniques, such as A* and DWA, were used for planning and obstacle avoidance.
Over time, tilting 2D laser scanners have given way to ubiquitous inexpensive 3D depth cameras and laser scanners.
Some of the approach's formulation is specialized for the geometry, low accuracy, and effects displayed by a 2D tilting laser configuration.
Additionally, with the introduction of sparse multi-beam laser scanners, pure raycasting methods can no longer effectively clear out free space, especially in a dynamic scene.
STVL created a more scalable alternative for many, sparse, and long-range sensors that is utilized in place of such techniques \cite{stvl}.

An outdoor robot navigation experiment was presented in 2013 \cite{urban_nav}.
That work focused on the mapping, localization, and simple dynamic obstacle detection method used to navigate the streets of Germany over 7 km.
Our work focuses on a modular navigation framework to build many types of applications using behavior trees and extended experimentation to showcase its maturity.
While we do not directly consider the detection of dynamic obstacles in our experiments, all the necessary interfaces exist to do this through Navigation2 with an use-case specific detector enabled.
We implemented an optimal local trajectory planner and 3D environmental modelling technique that is functionally efficient in the presence of dynamic obstacles without explicit detection, but can be augmented with such information if available. 

\subsection{Models}
Behavior trees (BT) have been shown to be successful for task planning in robotics, as a replacement for finite state machines (FSM) \cite{soccer}.
With growing numbers of states and transitions, modeling complex behaviors and multi-step tasks as FSMs can become intractably complex for real-world tasks. 
\cite{soccer} successfully demonstrates the use of behavior trees to model motion and shooting tactics for soccer game play that would otherwise be intractable with a FSM.

BTs have also been applied heavily to the domain of robot mission execution, robot manipulation, and complex mobile-manipulator tasks \cite{bt_intro}.
Further, behavior trees can be easily modified with widely reusable primitives and loaded at run-time, requiring no programmed logic.
For this reason, behavior trees are used in our approach as the primary structure of the navigation framework.

%% file: body/design.tex
\section{Navigation2 Design}
\label{sec:design}

The Navigation2 system was designed for a high-degree of configurability and future expansion.
While the marathon experiments display one such configuration of Navigation2, the intent of our work is to support a large variety of robot types (differential, holonomic, ackermann, legged) for a large variety of environments and applications.
Further, this support is not only targeted for research and education, but for production robots as well.
The design took into account the requirements for robotics products needs including safety, security, and determinism without loss of the above generality.

\subsection{Reliable}
\label{subsec:reliable}

For professional mobile robots moving in excess of 2 m/s near humans, there are many safety and reliability concerns that must be addressed.
Navigation2 is designed on top of a real-time capable meta-operating system, ROS2, to address functional safety standards and determinism.
ROS2 is the second generation of ROS, the popular robot middleware, built on Data Distribution Service (DDS) communication standard \cite{ros2_performance}.
DDS is used in critical infrastructure including aircraft, missile systems, and financial systems.
It exposes the same strict messaging guarantees from the DDS standard to enable applications to select a policy for messaging.
ROS2 leverages DDS security features to allow a user to securely transmit information inside the robot and to cloud servers without a dedicated network.
These features create a viable communication framework for industrial-grade mobile robots.

Further, ROS2 introduces the concept of Managed Nodes (also known as Lifecycle Nodes). 
A managed node implements a server structure with clear state transitions from instantiation through destruction \cite{managed_nodes}.
This server is created when the program is launched, but waits for external stimulus to transition through a deterministic bringup process.
At shutdown or error, the server is stepped through its state machine from an active to finalized state.
Each state transition has clear responsibilities including management of memory allocation, networking interfaces, and beginning to process its task.
All servers in Navigation2 make use of managed nodes for deterministic program lifecycle management and memory allocations.

\subsection{Modular and Reconfigurable}
\label{subsec:modular}

To create a navigation system that can work with a large variety of robots in many environments, Navigation2 must be highly modular.
Similarly, our design approach favored solutions that were not simply modular, but also easy to reconfigure and select at run-time.
Navigation2 created two design patterns to accomplish this: a behavior tree navigator and task-specific asynchronous servers, shown in Figure \ref{fig:architecture}.
Each task-specific server is a ROS2 node hosting algorithm plugins, which are libraries dynamically loaded at run-time.

\input{figs/diagrams/architecture}

The \textit{Behavior Tree Navigator} uses a behavior tree to orchestrate the navigation tasks; it activates and tracks progress of planner, controller, and recovery servers to navigate. 
The behavior tree plugins call to the planner, controller and recovery servers using the ROS2 action interface.
Unique navigation behaviors can be created by modifying a behavior tree, stored as an XML.
BTs are trivial to reconfigure with different control flow and condition node types, requiring no programming.
Changing navigation logic is as simple as creating a new behavior tree markup file that is loaded at run-time.
We exploit BT action node statuses and control flow nodes to create a contextual recovery system where the failure of a specific server will trigger a unique response. 

Using ROS2, the behavior tree nodes in the navigator can call long running asynchronous servers in other processor cores. 
Making use of multi-core processors substantially increases the amount of compute resources a navigation system can effectively utilize.
The Navigation2 BT navigator makes use of dynamic libraries to load plugins of BT nodes.
This pattern allows for reusable primitive nodes to be created and loaded with a behavior tree XML at run-time without linking to the navigator itself.
More than simply utilizing multiple processor cores, these nodes can call remote servers on other CPUs in any language with ROS2 client library support.

The task-specific servers are designed to each host a ROS2 server, environmental model, and run-time selected algorithm plugin.
They are modular such that none or many of them can run at the same time to compute actions.
The ROS2 server is the entry-point for BT navigator nodes.
This server also processes cancellation, preemption, or new information requests.
The requests are forwarded to the algorithm plugin to complete their task with access to the environmental model.

In summary, we can create configurable behavior trees whose structure and nodes can be loaded at run-time.
Each BT node manages the communications with a remote server, that hosts an algorithm written in an array of languages, loaded at run-time through plugin interfaces. 

\subsection{Support Feature Extensions}

Using the modularity and configurability of Navigation2, several commercial feature extensions were proposed in the design phase.
This subsection will highlight key examples and the design considerations to enable them.
Behavior trees are most frequently used to model complex or multi-step tasks, where navigating to a position is a node of that larger task.
The BT library \textit{BehaviorTree.CPP}\footnote{\url{https://github.com/BehaviorTree/BehaviorTree.CPP}} was selected for its popularity and support for subtrees to allow Navigation2 to be used in larger BT tasks.
Users with complex missions may use a provided 'Navigation2 BT node' wrapper to use Navigation2 as a subtree of their mission.

Many service robots will autonomously dock to recharge or operate elevators without requiring human assistance \cite{savioke}.
Traditional navigation frameworks have not addressed this class of task in their formulation.
While these extensions are not implemented due to lack of vendor standardization, we provide instruction to easily enable them.
Each server supports multiple algorithm plugins; a docking controller algorithm can be loaded into the controller server and called from the BT to dock a robot.
A BT action node can be trivially added to call an elevator using other IoT APIs provided by vendors \cite{kone}.

%% file: figs/diagrams/architecture.tex
\begin{figure}[ht]
    \centering
    \includegraphics[
        width=0.48\textwidth]
        {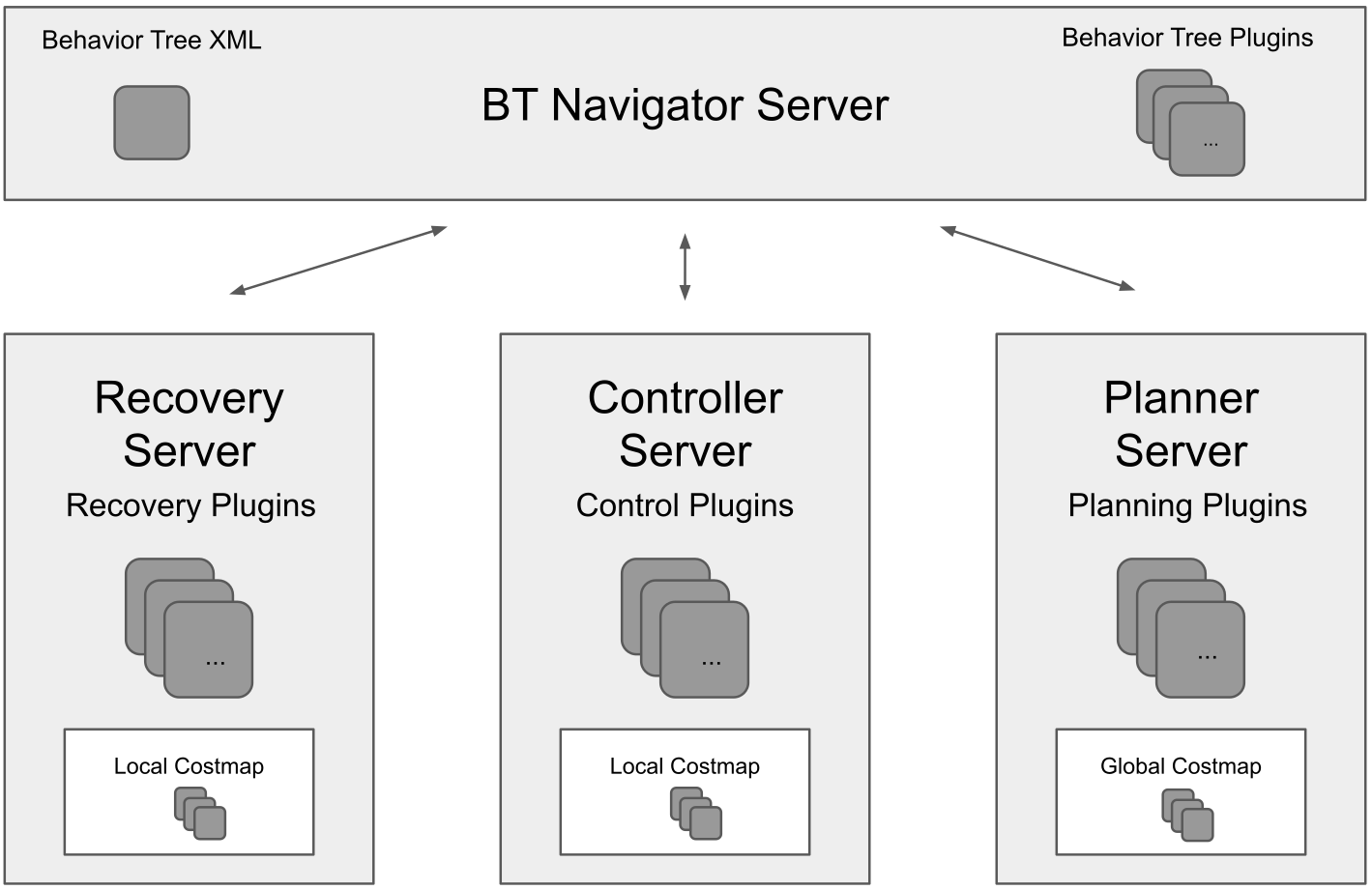}
    \caption{Overview of Navigation2 design.}
    \label{fig:architecture}
\end{figure}

%% file: body/implementation.tex
\section{Navigation2 Implementation}
\label{sec:implementation}

In this section, each of the recovery, planner, and controller servers provided in Navigation2 are examined.
Each may be customized or replaced by the user, but we will address the specific algorithm plugins and servers provided, recommended, and used in our experiment.
Each server contains an environmental model relevant for their operation, a network interface to ROS2, and a set of algorithm plugins to be defined at run-time. Provided plugins can be seen in Table \ref{tab:navigation_plugins}.


\input{figs/tables/navigation_plugins}


\subsection{Behavior Tree Navigator}

The behavior tree navigator is the highest level component of Navigation2, hosting the tree used to implement navigation behaviors. 
Navigation2 builds upon BehaviorTree.CPP, an open source C++ library supporting type-safe asynchronous actions, composable trees, and logging/profiling infrastructures for development.
The top level node, BT navigator server, seen in Figure \ref{fig:architecture}, is where user defined BT are loaded and run. Each node calls a server, below, to complete a task.

\input{figs/diagrams/behavior_tree}

Figure \ref{fig:behavior_tree} depicts a conventional Navigation2 BT, also used in our experiments.
Following the logical flow from left to right, a policy ticks the global planner action at a rate of 1 Hz.
If the global planner fails to find a path, the fallback node advances to the next child action to clear the global environment of prior obstacles.
Clearing of the environmental model may help resolve potential failures in the perception system.
The parent sequence node then ticks the controller or similar clear fallback behavior.
Should local fallback actions and subsequent sequence node return failure, the root fallback node then ticks the final child, attempting evermore aggressive recovery behaviors.
First by cleaning all environments, then spinning the robot in place to reorient local obstacles, and finally waiting for any dynamic obstacles to give way. 


\subsection{Recovery}

Recovery behaviors are used to mitigate complete navigation failures. They are called from the leaves in the behavior tree and carried out by the recovery server.
By convention, these behaviors are ordered from conservative to aggressive actions.

Note, however, that recovery behaviors can either be specific to its subtree (e.g. the global planner or controller) or system level in the subtree containing only recoveries in case of system failures.
Those used for experiments are as follows:

\textbf{Clear Costmap}: A recovery to clear costmap layers in case of perception system failure.
\textbf{Spin}: A recovery to clear out free space and nudge robot out of potential local failures, e.g. the robot perceives itself to be too entrapped to back out.
\textbf{Wait}: A recovery to wait in case of time-based obstacle like human traffic or collecting more sensor data.

\subsection{Perception}

As range and resolution of depth sensing technologies have improved, so too have world representations for modeling a robot's environment.
To leverage these improvements, a layered costmap approach is used, allowing the user to customize the hierarchy of layers from various sensor modalities, resolutions, and rate limits.
A layered costmap allows for a single costmap to be coherently updated by a number of data sources and algorithms.
Each layer can extend or modify the costmap it inherits, then forward the new information to planners and controllers.

\begin{figure}[ht]
\centering
    \includegraphics[
        width=0.9\linewidth,
        ]
        {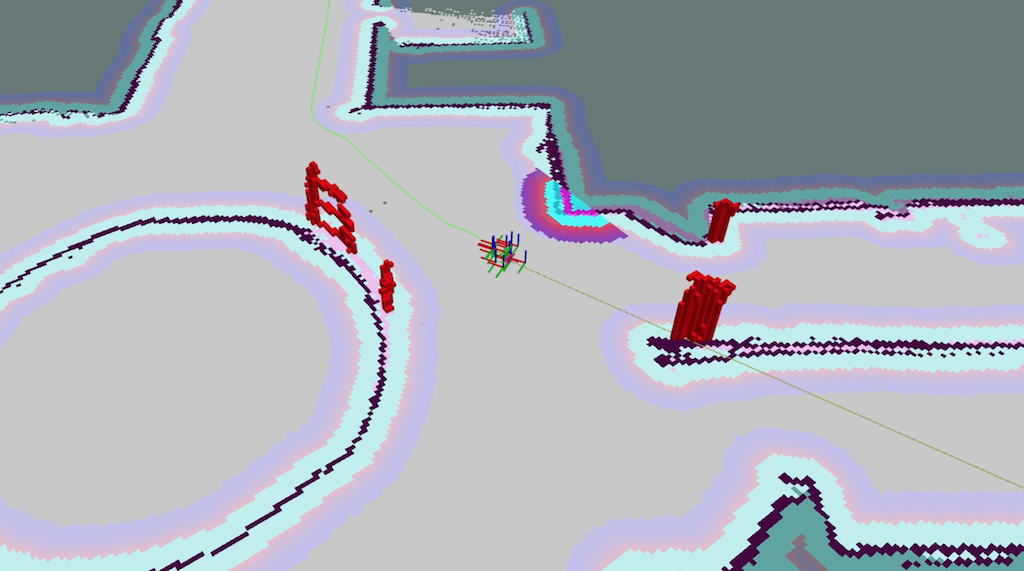}\\
    \vspace{0.2cm}
    \includegraphics[
        width=0.9\linewidth,
        ]
        {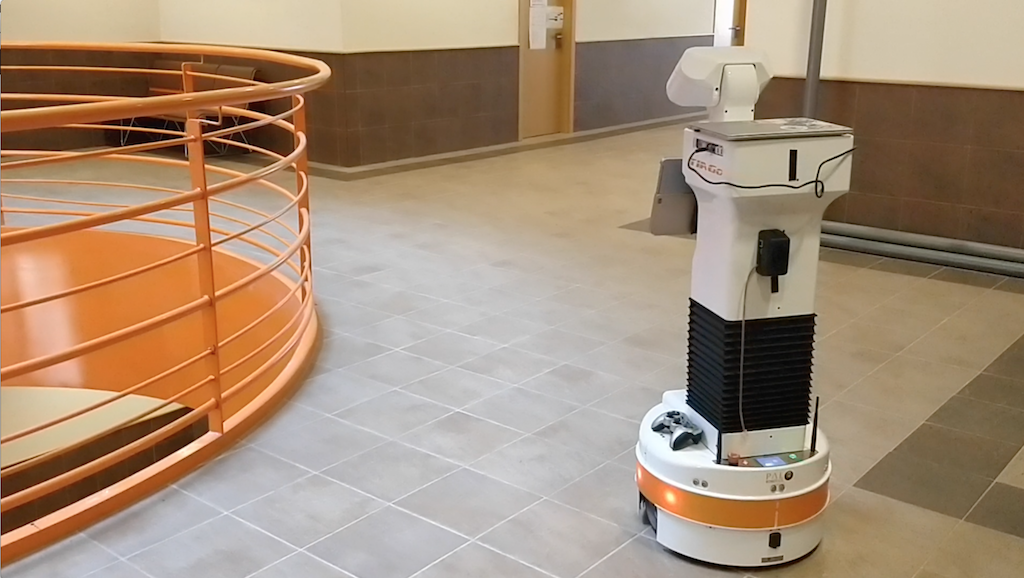}
    \caption{Visualization of robot with depth, and map information.}
    \label{fig:perception}
\end{figure}

While the costmap is two dimensional, STVL maintains a 3D representation of the environment to project obstacles into the planning space.
This layer uses temporal-based measurement persistence to maintain an accurate view of the world in the presence of dynamic obstacles.
It scales better than raycasting approaches with many, high-resolution, and long-range sensors such as 2D and 3D laser scanners, depth cameras, and Radars \cite{stvl}.
Figure \ref{fig:perception} shows the robot planning a route in a central stairwell and its view of the environment using STVL to represent a voxel grid. The map (shown in black) and inflation of the static map (colored gradient) are also displayed in the local and global costmaps.

Those used for experiments are as follows:

\textbf{Static Layer}: Uses the static map, provided by a SLAM pipeline or loaded from disk, and initializes occupancy information.
\textbf{Inflation Layer}: Inflates lethal obstacles in costmap with exponential decay by convolving the collision footprint of the robot.
\textbf{Spatio-Temporal Voxel Layer}: Maintains temporal 3D sparse volumetric voxel grid that decays over time via sensor models from the laser and RGBD cameras.

\subsection{Global Planner and Controller}

The task of the global planner is to compute the shortest route to a goal and the controller uses local information to compute the best local path and control signals.
The global planner and controller are plugins delegated to task-specific asynchronous servers.
For performance, relevant global and local costmaps are co-located in their respective server.

New algorithms are employed for smoothly navigating in highly dynamic environments by combining the TEB Local Planners with the STVL for dynamic 3D perception.
The TEB controller is capable of taking in object detections and tracks to inform the controller of additional constraints on the environment.
It is also suitable for use on differential, omnidirectional, and ackermann style robots.
DWB, a DWA implementation (DW "B" being incrementally named), is also implemented as a highly-configurable scoring-based controller.
However it is not used in these experiments due to its performance in dynamic scenes.
Those used for experiments are as follows:

\textbf{A* Planner}: Navigation function planner using A* expansion and assumes a 2D holonomic particle.
\textbf{TEB Controller}: Uses Timed-elastic-bands for time-optimal point-to-point nonlinear model predictive control.

\subsection{State Estimation}

For state estimation, Navigation2 follows ROS transformation tree standards to make use of many modern tools available from the community.
This includes \textbf{Robot Localization}, a general sensor fusion solution using Extended or Unscented Kalman Filters  \cite{rl}.
It is used to provide smoothed base odometry from \textit{N} arbitrary sources, which often includes wheel odometry, multiple IMUs, and visual odometry algorithms.

However, locally filtered poses are insufficient to account for integrated odometric drift, thus a global localization solution remains necessary for navigation.
The global localization solutions used for experiments are as follows:

\textbf{SLAM Toolbox}: A configurable graph-based SLAM system using 2D pose graphs and canonical scan matching to generate a map and serialized files for multi-session mapping \cite{slam_toolbox}.
This was used to create the static map in preparation for the experiment.
\textbf{AMCL}: An implementation of an Adaptive Monte-Carlo Localization, which uses a particle filter to localize a robot in a given occupancy grid using omni-directional or differential motion models \cite{Fox2002}. This was used as the primary localization tool for the duration of the experiments.

\subsection{Utilities}

Navigation2 also includes tools to aid in testing and operations.
It includes the \textit{Lifecycle Manager} to coordinate the program lifecycle of the navigator and various servers.
This manager will step each server through the managed node lifecycle: inactive, active, and finalized.

The lifecycle manager can be set to transition all provided servers to the active state on system bringup.
Alternatively, it may wait for a signal to activate the system.
Navigation2 enables users to activate, manage, and command their system through a GUI.
This GUI includes controlling the lifecycle manager, issuing navigation commands, selecting waypoints to navigate through, and canceling current tasks in real-time.
For autonomous applications, all of these commands are available through ROS2 server interfaces with examples.

\subsection{Quality Assurance}

System wide integration tests are employed for quality assurance purposes; helping to avoid regressions over time.
In addition to basic unit testing, code style linters, and memory static analysis - simulation tests are used by a continuous integration pipeline to emulate an entire robot software stack.
As shown in Figure \ref{fig:turtlebot_testing}, a 3D robotic simulator, Gazebo\cite{gazebo}, is used to test community contributions over an assortment of virtual scenarios including static and dynamic environments.

\input{figs/images/turtlebot_testing}

Such system tests allow maintainers to monitor not only navigation failures over tested trajectories, but also performance analytics for various algorithms over an assortment of robot platforms with unique locomotion kinematics.

%% file: figs/tables/navigation_plugins.tex
\begin{table}[ht]
\caption{\label{tab:navigation_plugins}Navigation2 provided plugins.}
\begin{tabular}{|l|l|l|}
\hline
\textbf{Type}                       & \textbf{Plugin}              & \textbf{Description} \\ \hline
\multirow{2}{*}{Control}   & DWB Controller      & Configurable DWA controller \\ \cline{2-3} 
                           & \textbf{TEB Controller}      & Timed-Elastic-Bands controller \\ \hline
\multirow{6}{*}{Costmap}   & \textbf{Inflation Layer}     & Inflate obstacles in costmap \\ \cline{2-3} 
                           & Non-Persist. Voxel  & Maintain only recent voxels \\ \cline{2-3} 
                           & Obstacle Layer      & Raycast 2D obstacles \\ \cline{2-3} 
                           & \textbf{STVL}   & Temporal 3D sparse voxel grid \\ \cline{2-3} 
                           & \textbf{Static Layer}        & Loads static map into costmap \\ \cline{2-3} 
                           & Voxel Layer         & Raycast 3D obstacles \\ \hline
\multirow{1}{*}{Planner}   & \textbf{NavFn Planner}       & Holonomic A* expansion \\ \hline
\multirow{4}{*}{Recovery}  & Back Up             & Back out of sticky situations \\ \cline{2-3} 
                           & \textbf{Clear Costmap}       & Clear erroneous measurements \\ \cline{2-3} 
                           & \textbf{Spin}                & Rotate to clear free space \\ \cline{2-3} 
                           & \textbf{Wait}                & Waitout time based obstacles \\ \cline{2-3} 
\hline
\end{tabular}
\centering \\
Bold entries used in experiments.

\end{table}

%% file: figs/diagrams/behavior_tree.tex
\begin{figure}[ht]
    \centering
    \includegraphics[
        width=0.48\textwidth]
        {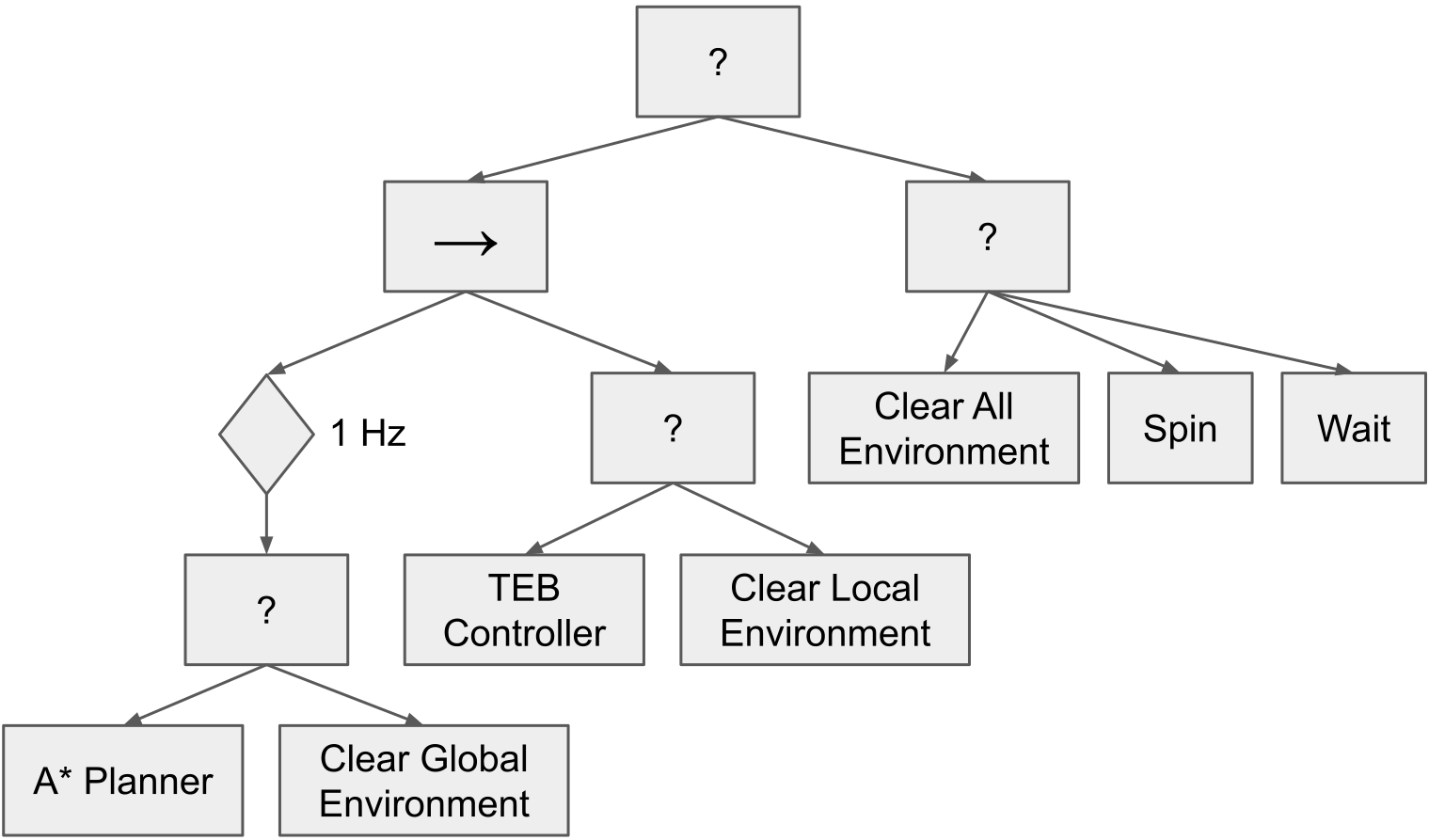}
    \caption{Behavior tree used in the marathon. '?' represent a fallback node, and "$\to$" represents a sequence node.}
    \label{fig:behavior_tree}
\end{figure}

%% file: figs/images/turtlebot_testing.tex
\begin{figure}[ht]
    \centering
        \includegraphics[page=1,
            width=0.9\linewidth]
            {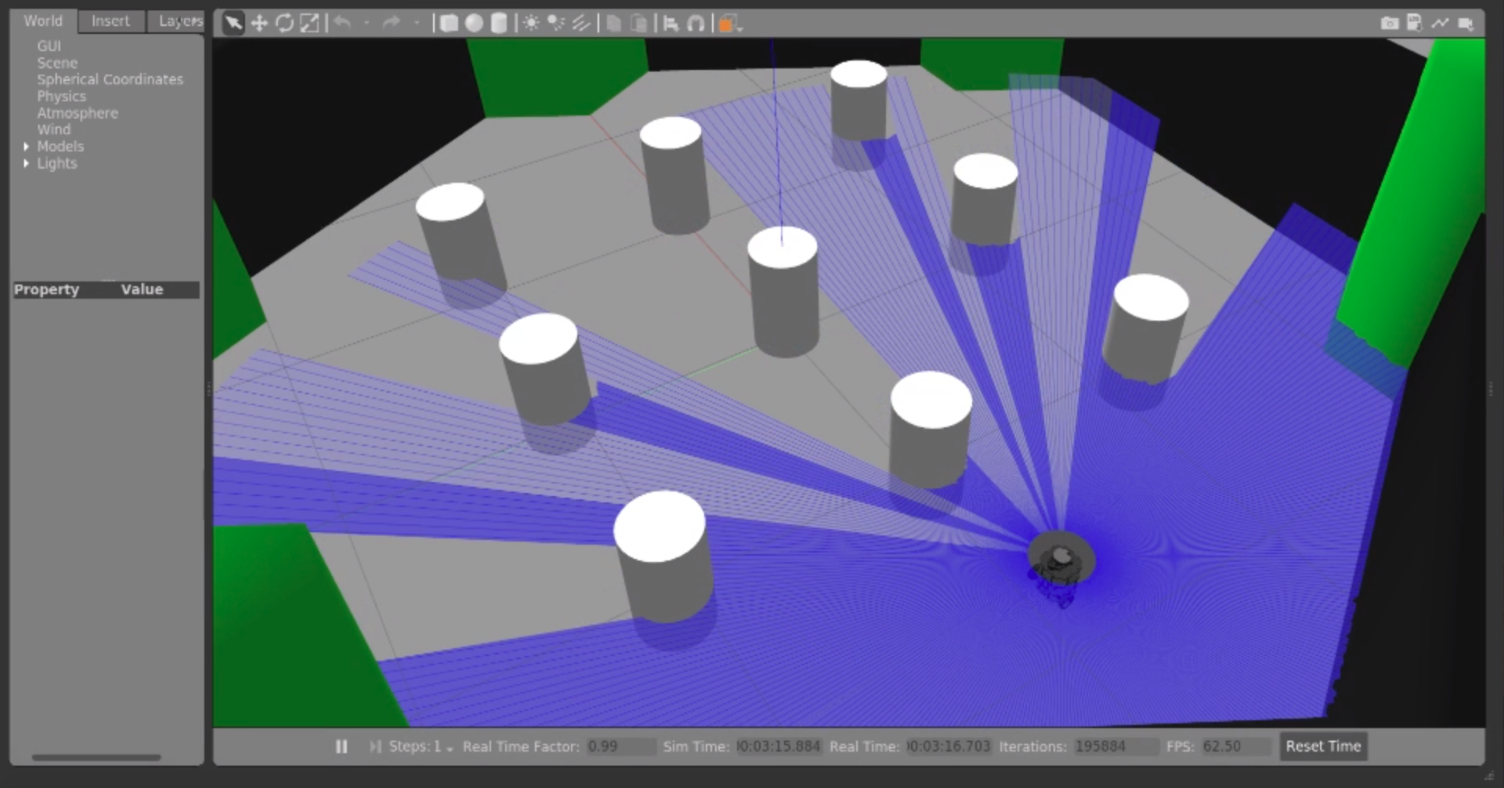}
    \caption{Sample simulation from Continuous Integration tests.}
    \label{fig:turtlebot_testing}
\end{figure}

%% file: body/analysis.tex
\section{Experiment and Analysis}

\subsection{Experiment Overview}

To show the robustness of Navigation2, we conducted a more extensive experiment than proposed in \cite{eppstein}, an "ultra-marathon" (a distance greater than a standard marathon).
This test demonstrated the robustness and reliability of our system in long-term operations without human assistance using two professional robots: the Tiago and RB-1 base, shown in Figure \ref{fig:robots}.
These robots operated in a human-filled environment in a University setting running at near-industrial speeds.

\input{figs/images/experiment_robots}

The experiment took place in the Technical School of Telecommunications Engineers at the Rey Juan Carlos University.
Figure \ref{fig:map_hallway} shows the map of this environment with the track used in the experiment.
The robot was made to navigate this track around a central stairwell (surrounding waypoints 3, 11, 13, 14), a high-traffic bridge and hallway (waypoints 4-10), and back into a lab (waypoint 1) through a narrow doorway without intervention. 
The robot shared the environment alongside students, Figure \ref{fig:exps_robots}, whom sometimes spontaneously block the path of the robot.
We define a 300 meters route for the experiment through this set of waypoints. 

\input{figs/images/map_hallway}

\input{figs/tables/robots}

The characteristics of both robots are similar, described in Table \ref{tab:robots}.
Both use differential steering, with similar diameters but varying heights.
They each use RGBD cameras and safety laser scanners with different ranges and resolutions for perception and obstacle avoidance.
Both robots were set to have a maximum speed of 0.45 m/s for functional safety, below their maximum allowable speed.
Two different robots were used to show the portability of our system across multiple hardware vendors with trivial reconfiguration.
The Tiago utilized an external computer because its internal operating system is not supported by ROS2.
Both robot vendors provide supported drivers and interfaces for users.
We created a custom bridge between it and ROS2 which supports continuous flow of data.
This was required to support high-frequency data requirements on transformations and odometry.

The robot saves the following information once a second:

\begin{itemize}
    \item \texttt{timestamp}: Current time.
    \item \texttt{distance}: Odometric distance travelled.
    \item \texttt{recovery\_executed}: Logged executed recoveries.
    \item \texttt{vel\_x}: Instantaneous linear velocity.
    \item \texttt{vel\_theta}: Instantaneous angular velocity.
\end{itemize}

All of the software, configurations, and data used in this experiment is available in the marathon\footnote{https://github.com/IntelligentRoboticsLabs/marathon\_ros2} repository.
This repository contains instructions to reproduce the experiment as well as the data described above collected in the experiment.
We collected the following information for the experiment dataset: localized pose with covariance, instantaneous speed commanded, whether a recovery behavior was triggered, time and distance navigated, laser scan readings, geometric transformations, inertial sensor data, and map. This data is publicly available in the repository for experimental verification.

\subsection{Analysis}

\input{figs/tables/exp_results}

Table \ref{tab:exp_results} shows the results of the experiment.
The robots successfully navigated over 37 miles in under 23 hours in a dynamic campus environment.
During the experiment the average linear speed was 0.37 m/s.
Neither robot suffered a collision or a dangerous situation requiring an emergency stop during the test.
The robots operated without direct assistance throughout the experiment and never failed to complete a navigation task.

\begin{figure*}[ht]
    \centering
    \includegraphics[
        width=0.98\textwidth,
        ]{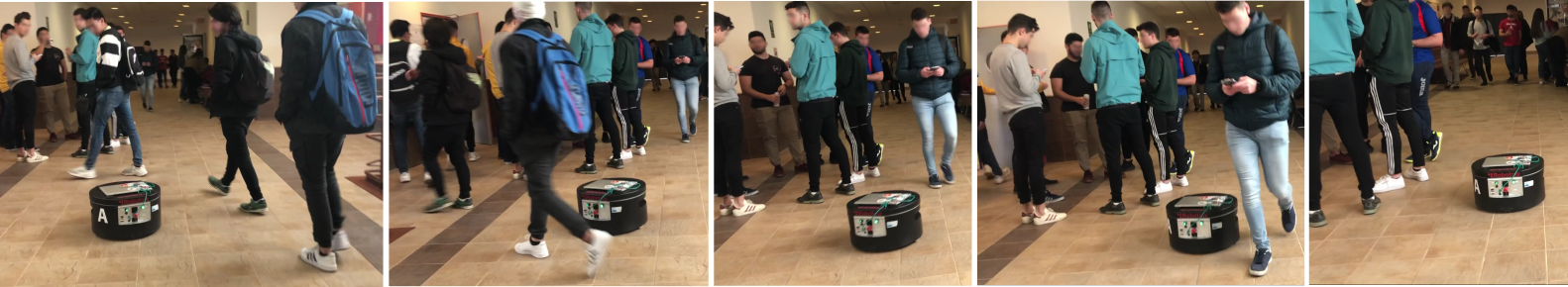}
    \caption{An example robot-pedestrian interaction.}
    \label{fig:panel}
\end{figure*}

Figure \ref{fig:panel} shows a time-lapse panel of a situation that the robot faced in a crowded hallway during the moments of the start of classes.
In this period, many students are walking quickly to their destinations, frequently using phones.
This sequence shows how the robot interacts with several groups of students.
On approach, the robot first slows and navigates around first two people in the group in the first panel.
The third person pauses, giving the robot the right of way to continue -- in panel 2.
Then, after navigating around the first group in panels 1-3, a new person from panel 3 to 4 walks through the scene and the robot corrects its trajectory to navigate away from their path.
Panel 5 shows the robot exiting the complex scenario without collision or stoppage.

Passive assistance was rendered to the robot on rare instances when students occupied the exact pose of a waypoint the robot was navigating towards.
At that time, students were asked to move to allow the robot to continue.
The time impact of this passive assistance was under 1 minute over the 22.8 hour experiment.
This was a fault in the application built for the marathon experiment, which utilizes Navigation2.
The application did not account for occupation of the goal pose, however, this does not represent an issue or failure mode of Navigation2 itself.

While the total number of recoveries seems high, triggering recovery behaviors is not generally a poor-performance indicator since it is part of a fault-tolerant system.
Recoveries indicated that the robot had some difficulty due to a potentially blocked path or erroneous sensor measurement.
The system automatically performs the recovery required to overcome a given challenge resulting in the robust system presented.

The majority of recovery behaviors executed were due to two cases.
The first happened in crowded spaces where the robot was unable to compute a route to its goal.
In this case, the robot cleared its environmental representation, usually the local costmap, of obstacles.
If this was insufficient, typically, the path was blocked by students and the wait recovery paused navigation until the path was clear.
The second case was when the localization confidence was low.
Long corridor areas with repetitive features in the presence of many people around caused occasional issues.
AMCL works well in dynamic environments when enough rays can reach the static map, however, many changes over a long duration can degrade positioning quality.
In this situation, the robot's spin recovery helped regain localization confidence to continue on its path.

A video highlight reel of this experiment can be found in the footnote \footnote{\url{https://www.youtube.com/watch?v=tXp8wFZr68M}}.

%% file: figs/images/experiment_robots.tex
\begin{figure}[t]
\centering
    \includegraphics[
        width=0.9\linewidth,
        trim= 300 100 0 100,
        clip
        ]
        {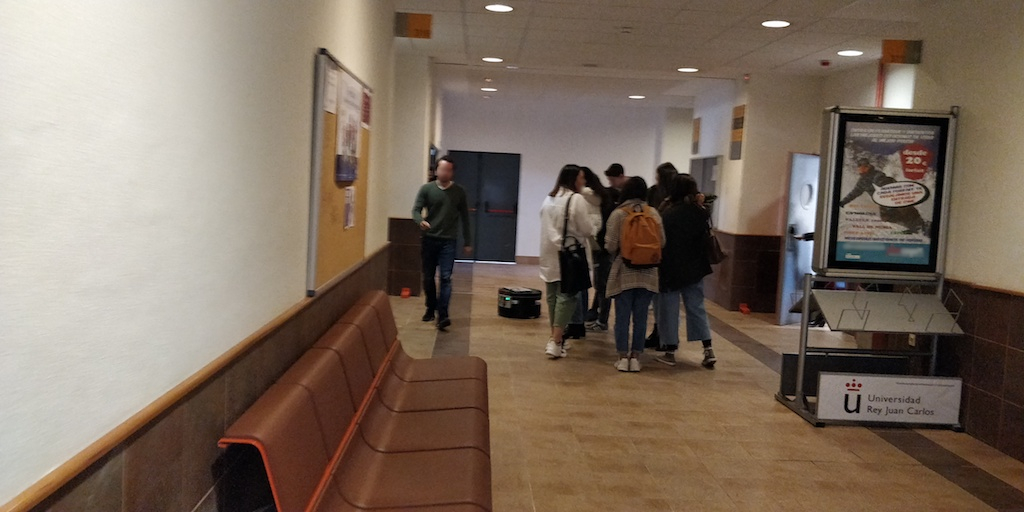}\\
    \vspace{0.1cm}
    \includegraphics[
        width=0.9\linewidth,
        trim= 50 100 0 25,
        clip
        ]
        {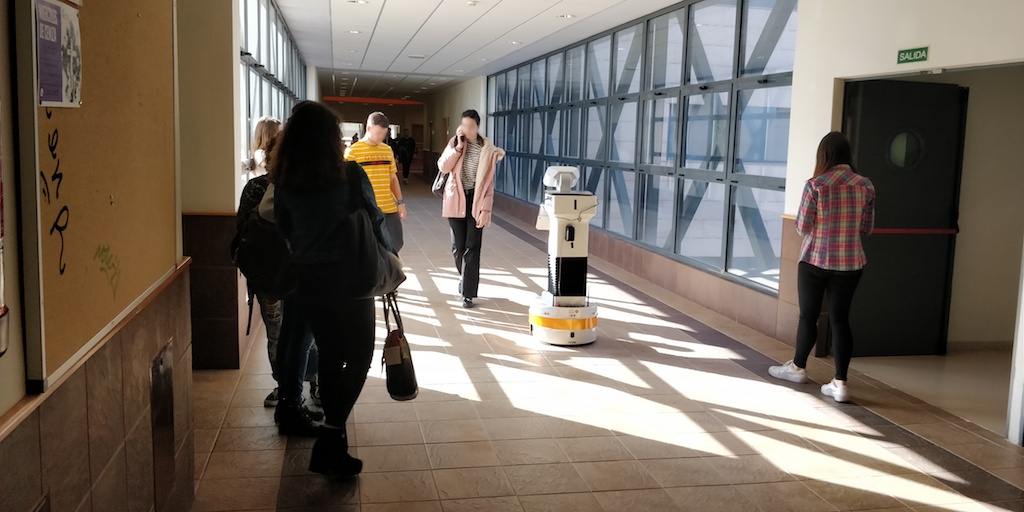}
    \caption{Robots during the experiment.}
    \label{fig:exps_robots}
\end{figure}

%% file: figs/images/map_hallway.tex
\begin{figure*}[!htb]
    \centering
    \includegraphics[
        width=0.98\textwidth]
        {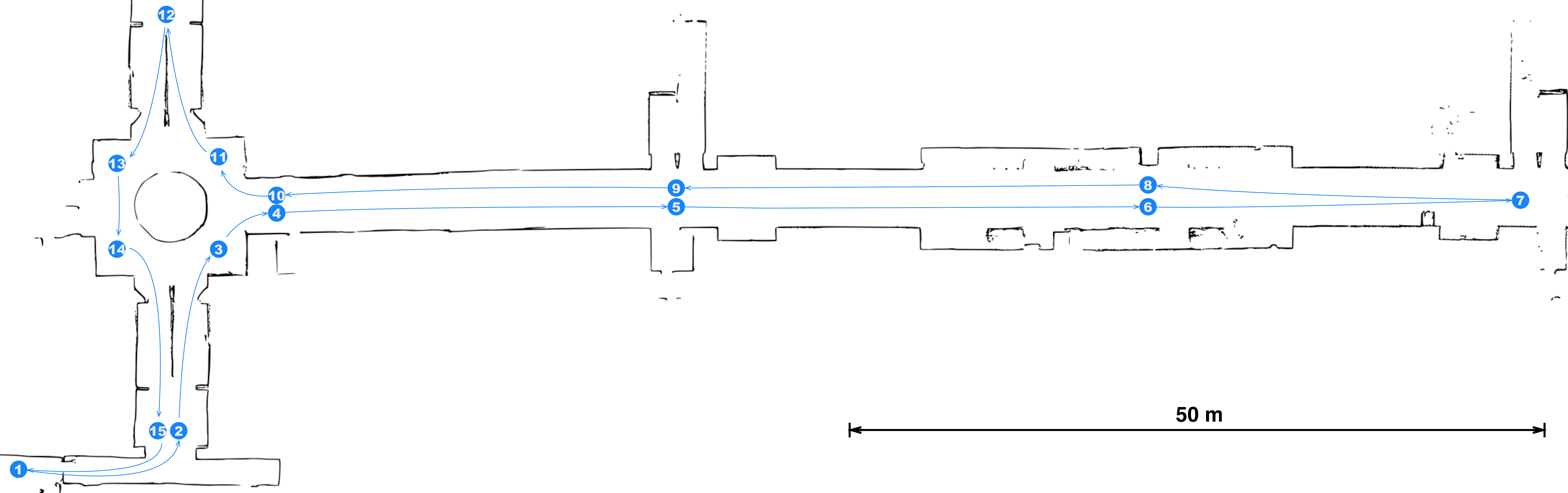}
    \caption{Map of hallway, stairwell, and lab used in experiments.}
    \label{fig:map_hallway}
\end{figure*}

%% file: figs/tables/robots.tex
\begin{table}[!htb]
\caption{\label{tab:robots}Robot specifications.}
\centering

\begin{tabular}{|c|c|c|}
\hline
\textbf{}                   & \textbf{Tiago} & \textbf{RB-1}           \\ \hline
\textbf{Robot Type}         & Base + Torso                             & Base                                 \\ \hline
\textbf{Dimensions}         & 0.54 m x 1.45 m          & 0.50 m x 0.251 m        \\ \hline
\textbf{Max. Speed}         & 1.0 m/s                                 & 1.5 m/s                              \\ \hline
\textbf{Battery}            & 2 x 36V 20Ah                     & 24V 30Ah                  \\ \hline
\textbf{Compute}            & Intel i7-7700              & Intel i7-7567U              \\ \hline
\textbf{Memory}            & 8GB Ram/250GB SSD              & 8GB Ram/120GB SSD            \\ \hline
\textbf{Operating System}            & Ubuntu 16.04              & Ubuntu 18.04            \\ \hline
\textbf{Depth Camera}       & Orbbec Astra S     & Orbbec Astra \\ \hline
\textbf{Laser Sensor}       & Sick TIM561    & Sick Tim571          \\ \hline
\textbf{Navigation2}        & External             & Onboard    \\ \hline
\end{tabular}

\end{table}

%% file: figs/tables/exp_results.tex
\begin{table}[ht]
\caption{\label{tab:exp_results}Experiment results for both robots.}
\centering
\begin{tabular}{|c|c|c|c|}
\hline
                    & \textbf{RB1}    & \textbf{TIAGo} & \textbf{Total} \\ \hline
\textbf{Time (hrs)}        & 9.4    & 13.4  & \textbf{22.8} \\ \hline
\textbf{Distance (miles)}    & 15.6   & 21.8  &  \textbf{37.4}\\ \hline
\textbf{Recoveries}          & 52     & 116   & \textbf{168}  \\ \hline
\textbf{Recoveries per mile} & 3.3    & 5.3   & \textbf{4.3}  \\ \hline
\textbf{Avg. speed (m/s)}    & 0.39   & 0.35  & \textbf{0.37} \\ \hline
\textbf{Max speed (m/s)}     & 0.45   & 0.45  & \textbf{0.45}  \\ \hline
\textbf{Num. of Collisions}          & 0      & 0     & \textbf{0} \\ \hline
\textbf{Num. of Emergency stops}     & 0      & 0     & \textbf{0}  \\ \hline
\end{tabular}

\end{table}

%% file: body/limitations.tex
\section{Limitations}
\label{sec:limitations}

While shown in experiments to effectively navigate in narrow spaces around people for many miles without human intervention, there are some limitations to the Navigation2 system which will be addressed in future work.
Currently, the A* planner used does not create feasible paths for non-circular non-holonomic robots.
This does not generally cause an issue for differential drive robots used in this experiment as they can rotate to a new heading in-place.
However, this creates an issue for complex geometry robots or car-like robots resulting in an inability to move because of an infeasible path  to follow.
Extensions are in development to support non-holonomic robots of arbitrary shape.

Further, the TEB controller can account for dynamic obstacles in computing velocity commands, however, no explicit obstacle detection was utilized in these experiments.
Without these explicit detections, our system was able to successfully navigate in the presence of many dynamic obstacles.
However, detections and predictive models will further enable robots to operate safely around humans and other agents.

%% file: body/conclusion.tex
\section{Conclusion}
\label{sec:conclusion}

It has been shown that the Navigation2 navigation system can reliably navigate a campus environment in the presence of students during long-duration testing.
Our demonstration had 2 different industrial-grade robots, using our system, navigate in excess of a marathon without any human intervention or collision.
The system made use of behavior trees to orchestrate navigation algorithms to be highly configurable and leverage multi-core processors -- enabled by ROS2's industrial-grade reliable communications.
We used modern algorithms such as STVL and TEB for perception and control in large and dynamic environments while continuing to build on the legacy of the popular work of the ROS Navigation Stack.

There are gaps in dynamic obstacle tracking and planning which are being developed and will be addressed in future work.
This work is open-source under permissive licensing and we anticipate many more extensions and algorithms to be created using our framework, further improving over time.